# Artificial Neuron Modelling Based on Wave Shape

Kieran Greer, Distributed Computing Systems, Belfast, UK.
http://distributedcomputingsystems.co.uk
Version 1.2

***Abstract –*** This paper describes a new model for an artificial neural network processing unit or neuron. It is slightly different to a traditional feedforward network by the fact that it favours a mechanism of trying to match the wave-like 'shape' of the input with the shape of the output against specific value error corrections. The expectation is then that a best fit shape can be transposed into the desired output values more easily. This allows for notions of reinforcement through resonance and also the construction of synapses.

## 1   Introduction

This paper[1] describes initial ideas for a new artificial neuron, to be used as part of a neural network that relies on shape and resonance for its construction. Resonance has been used before, as part of an ART network [1], for example. This new neuronal unit however relies on the wave-like shape of the desired output and tries to match that as part of its construction process. It has a slightly different philosophy, where the neuronal cell tries to move away from an exact error correction process, to incorporate measuring similarities between the input and output nodes. This is based on the shape that is formed from the 'changes' between the output values. These changes then also relate directly to the input values that caused them. If the input data gets presented in a different order, then the related output data is altered by the same order, changing any related shape in a consistent way. It is really the differences between values that define what the function is. When learning a function, each input point is not rote learnt specifically, but the knowledge of the function is what to do over every possible input. This requires learning all of the points in-between the test dataset as well. It is also consistent with a traditional feedforward backprop algorithm [5]. That type of algorithm was also created from trying to differentiate over a continuous function.

---

[1] This paper was originally published on Scribd in April 2012. This version has been accepted by the journal 'BRAIN' (Broad Research in Artificial Intelligence and Neuroscience).

The philosophy of the new model is in line with previous publications ([2][3]) which note that human processing might be particularly good at measuring and comparisons, and also that duplication, to allow for comparisons, is important. If a wave-like shape is used, then as part of modelling a real biological neuron, the idea of resonance to reinforce this is possible. Waves can reinforce themselves when they are in sync with each other. This was also a reason for the ART neural network model, although that has a completely different architecture.

The rest of this paper is structured as follows: section 2 describes the model for the new neural network neuron. Section 3 gives a very basic overview of what algorithms might be used to create the network. Section 4 gives a proof of why the method should work. Section 5 gives an example scenario, while section 6 gives some conclusions on the work.

## 2   Neuron Model Architecture

The new model that is described here is attractive for several reasons. It does map to a real biological neuron quite closely. It allows for natural processes such as wave resonance and also allows for the idea of synapses that combine, in a more intelligent way, the potential set of inputs. Synapses are the links between the neuronal units, but they can carry more complex information than a single signal and they can join to create more complex networks. Traditional neural networks tend to send every output from one node to every node in the next level and rely on a different weight value for each input to code the network knowledge. This is clearly not the exact structure in a real human brain, where the output from one cell or neuron forms the input to only a certain number of other neurons. A typical processing unit in an artificial neural network might look as shown in Figure 1.

In Figure 1, each input *X* is weighted by a separate weight value *w*. These are then summed together to give a total input value for the neuron. This total can then be passed through a function, to transform it into the desired output value. This is then compared to the actual output value *d*; where the error or differences between the two sets is measured and used to correct the weight values, to bring the two sets of values closer. One way to update the weights is after each individual pattern is presented and processed. Another option is to process the error after the presentation of the whole dataset, as part of a batch update.

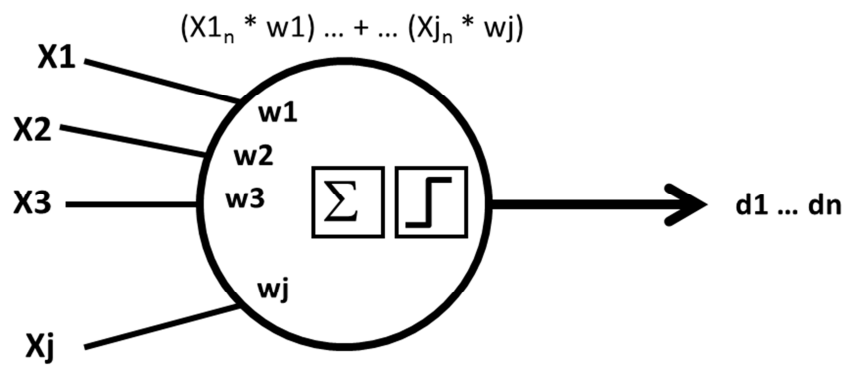

Figure 1. Typical feedforward artificial neural network processing unit.

For the new model, shown in Figure 2, the 'differences' between the outputs is measured and combined, to produce a kind of wave-like shape. If output point 1 has a value of 10, and output point 2 has a value of 5, for example, then this results in the shape moving down 5 points on the wave shape. This shape will, of course, change depending on the order that the dataset values are presented in. Also, if new data is presented, then that will produce a different shape. However, there is still an association between the input values and the output values and it is still this relation that is being learnt. The neural network needs to be able to learn a function that can generalise over the presented datasets, so that it can recognise the relation in previously unseen data as well. Generalising over trying to learn a wave-like shape, or trying to match the output values exactly looks quite similar, suggesting that the process is at least valid. In Figure 2, note that a transposition to move the learned shape up or down first is possible, before a weight value would try to scale it. If the initial match of the combined inputs can be as close as possible, then these adjustments will become less and will be more for fine tuning.

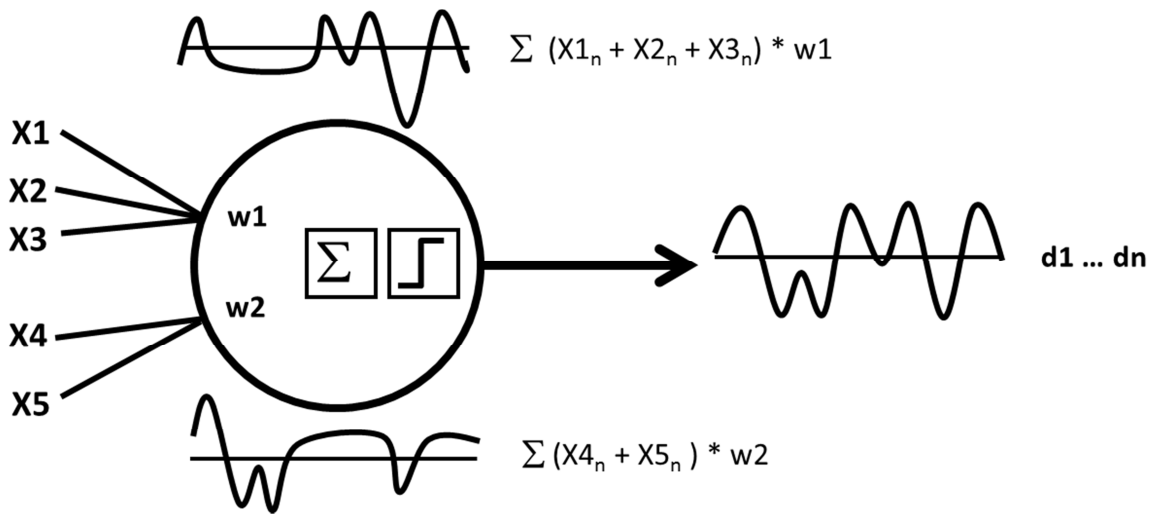

Figure 2. New neuronal model, based on matching a wave-like representation of the output.

It would probably be relatively easy to produce a 2-layer neural network based on this model. The error feedback is not particularly complicated and the update method is clear. The errors from the desired output pattern shapes adjust the weights, until the error is small enough to be tolerated. As described in section 3, this could even be performed in a single step, through averaging the total shape changes in the input and output datasets, to create a suitable weight value. If that process proves to be successful, then it will also be a very quick way to construct the neural network. The initial phase of forming the synapses will then become the main challenge.

Before the weight update process, a search process needs to try to match the inputs, to make their combined values as close as possible to the desired output and there can be more than one group of matches. The process would therefore need to recognise that two groups, with different results combined, would produce a shape closer to the desired goal. There could be technical reasons why this would be the case. If the shapes are as shown, then you can see that combining them gives a shape that is closer. Combining compensates where each is missing the correct shape part. Shape matching in hidden units is not so clear. A hidden unit would probably only want one output signal, if it was to keep the design integrity. Therefore, a hidden unit would not automatically be created and would receive input only from certain other nodes. It would then only link to certain output units along with other input units. Because the structure is flexible, an input to any neuron can be removed if it is not usefully matched to any group, thus reducing the number of connections in the final model.

## 3    Basic Algorithms

A basic form of algorithms that can be used to sort the inputs, find the best fit weight value and calculate the error, are listed in this section. These are described so that the process can be understood more easily, but they are not exact or final. The specific example in section 5 should also help. These algorithms are also more simplistic in the sense that they consider summing and averaging only. The wave shape is not considered in a more intelligent way. Therefore, changing any dataset order should not affect the result of these algorithms.

### 3.1    Group Inputs Together

The starting point for training the network is to find the best possible match of the input nodes to the output nodes. This is performed by looking at the whole dataset and trying to match over the dataset as follows:

- Determine the output shape by calculating the changes in the output values.
- Use a sort to try to group the inputs together to match this shape as closely as possible.
- Note that duplication of a shape by different sets of inputs would be allowed (even encouraged) and also different sets that would complement each other might also be allowed.

### 3.2    Adjust Vertically

Once the synapses (input groups) have been formed and their shape is clear, this can be compared to the output shape; with the goal of trying to centre both the output and the input shapes. This then makes the scaling through a weight adjustment easier. The vertical adjustment is made by comparing the overall shape magnitude, as follows:

- Measure the total input and the total output shape values and average for a single value.
- If the input is larger than the output, then move the input down to the level of the output.
- If the input is less than the output then move the input up to the level of the output.

### 3.3   Find Best Fit Weight Value

Once the shape of the input group is determined, it can be scaled with a single weight value. The process to find the best fit weight value is as follows:

- Measure the total input and the total output shape values and average for a single value.
- To try to match the output shape more closely, weight the input shape by the value *output shape change average / input shape change average*.
- It is possible that this can be performed after only one iteration, when further changes would not improve the shape.

### 3.4   Calculate the Actual Values Error

The error is then calculated for the differences between the actual and desired output values, not the shape. The vertical adjustment can also be included. As a final solution, each input group can only have a single weight value and this must be used for every data pattern that is presented:

- Get an average weight value for a synapse group from all of the test dataset values.
- The neuron actual output value is then all synapse input values, adjusted, and then scaled by their weight values.
- The error is the difference between this and the actual output value.

## 4   Proof

The following points describe how this model compares with a traditional feedforward network with a weight for each input. The error correction process for either model is to move the input value closer to the desired output value and is a converging process. If the input nodes in a traditional network give the same output values, then the error correction will change the weight for each one in the same way. It could therefore theoretically be replaced with a single weight value. As different input patterns will produce different value sets, this is not practically possible. To add robustness, a traditional neural network needs to seed the initial weight values randomly, so that it can adjust the error in different ways when looking for a solution. Instead of this, the new model combines inputs first, to get a better shape match. Each input group update could then be a single error correction, which is adjusted in much the same way as for a traditional network, only over different combined (synapse) value sets.

### 4.1  Robustness

Measuring the differences between the actual and the desired output 'shapes' is valid, because this measures a consistent change in the data. This is what associates the input with the related output value in a continuous and general way. So a single weight update for a 'group' of inputs is OK, but as the process then has less variability and generality (fewer weights), it should not work as well. One way to improve robustness is to have more than one input group with the correct shape. With different combinations, the sets of input values could still vary and so with new or previously unseen patterns, the behaviour of each group could still be different. Also for robustness, as input values are combined, an additional intelligent transformation takes place first. This gives an added dimension to the function that might compensate for the missing additional weights. Another test might be to randomly change the order of the presented data patterns, even for the same dataset. In some ways this is better than a traditional neural network, where the data points are more independent of each other, but then a real intelligence of the shape change is required. Because the input and output values will both be changed consistently for this, it should not result in a different function. There is therefore the idea of both horizontal and vertical shapes. When determining the synapse groups, this might be over a single (horizontal) pattern, or a consideration based on all input-output pairs. When considering or correcting the network error, it might be possible to produce a shape for each individual point, over all (vertical) datasets. Section 5 explains this further.

## 4.2 Generality

If the main process is shape matching, then this implies that most of the functions that are learnt by neural networks are able to match their shapes. Each input in a traditional neural network is assigned a weight value, but only one value for the whole dataset, which is also the case for the input synapse group of the new model. This weight therefore must either increase or decrease every value from that input in a consistent way, but cannot accommodate contradictory information. If the input presents the value 10 for an associated output value of 20 and also a value of -10 for an associated output value of 30, then a single weight value for that input will not be able to map both of these closely. The value of 2 is required to make the first input correct, while a value of -3 is required for the second one. This is again the case for either model. A traditional neural network can accommodate some level of redundancy, possibly of this sort, because other inputs will still be correct. The main design is still the same however and so there is not a great deal of difference between the two models in this respect. Initial tests over random data show that improvements in the initial input values are possible and sometimes substantial, but the proof is really intuitive. The most difficult task would probably be to find the best matching groups, when the weight scaling is then relatively straightforward. Note that the initial values (after sorting) can be used, without scaling, if no improvement can be found.

## 5  Example Scenario

This example scenario will hopefully help to describe the process further. The test dataset of Table 1 describes conditions under which sport gets played in a field somewhere. There are 4 weather input values and one play sport output value. There are two datasets in total. If the weather is good, that is high sun and light and low wind and rain, then lots of people play sport in the field. If the weather is poor however, that is low sun and light and high wind and rain, then very few people play sport in the field.

Table 1. Play Sport test dataset.

| Input / Output | Variable | Scenario 1 | Scenario 2 |
|---|---|---|---|
| **Input** | Sun | High | Low |
| **Input** | Daylight | High | Low |
| **Input** | Wind | Low | High |
| **Input** | Rain | Low | High |
| **Output** | Play Sport | High | Low |

The intended neural network might use this data as follows: the output shape is only the single high or low value, so group the sun and daylight input synapses together and the wind and rain ones. This is a horizontal shape match over each pattern and for more complex data, it might require intelligently considering lots of different shapes and combining into the most appropriate set for all patterns. The 'wind-rain' combination then requires a different weight value to reverse its input, to match the desired output.

When training or error correcting, consider vertical shape changes, where the 'sun-daylight' group moves from high to low, as does the desired output, while the 'wind-rain' group moves from low to high. Therefore, error corrections will adjust the two groups differently. If the 'sun-daylight' weight keeps the value as it is and the 'wind-rain' one reverses it, then if a new pattern has values of 'average' for all inputs, it would suggest an average value for play sport as well.

## 6  Conclusions

This paper has described a new model for an artificial neural network processing unit or neuron. It is based more on trying to match wave-like shapes than previous models have attempted. It also has relevance for modelling a real biological neuron, because resonance can play a part in the reinforcement mechanism and synapses can be formed. Even though the exact output is not matched, if the shape that represents the output value change is matched, then the input values can be converted into the correct output values more easily.

There is still a relation between the inputs and the outputs that causes this change. The adjustment might only require a vertical transposition and/or a scaling weight update. The process is more predictable than for a traditional neural network, as it does not use a random weight configuration setup. Unless the aggregation of input nodes is changed, there is no reason why the same dataset should produce different weight values. The network converges through a clearly defined process. There are however ways to introduce randomness into the learning process, to give it added robustness. One way would be the addition of different sets of hidden units that process inputs in a different ways. Another way is to randomly change the order of the presented data patterns, even for the same dataset. Or if this does not work, it is still an interesting result.

So the initial questions about the network would be accuracy, that is yet to be determined, and also its robustness or ability to generalise. The ability to generalise is assumed, because the process still maps the input to the output in a general way and tries to capture the continuous function that is determined by the shape. If the same input value is associated with a different output value in a different dataset, then a traditional neural network has the same problem.

While this is only a first step, the process does show promise. Future work will build a test system based on the new architecture and try to determine the levels of accuracy that can be achieved. There are many different types of test, based on the randomness and size of the data, but events have meant that it is easier to publish this new model now and try to provide a more comprehensive set of tests later. It would be interesting to try to find this matching wave-like shape in standard test datasets. Although this model has derived from the conclusions of the earlier work [2][3], on the surface it looks like a separate mechanism. The earlier work considered creating links, almost completely through feedback from the environment. There was no controlling algorithm that would try to force certain links only. This mechanism, on the other hand, does have a controlling algorithm that is likely to consistently produce the same result for the same dataset. Future work will also try to tie this network model in more with the whole cognitive process.